\documentclass[10pt,twocolumn,letterpaper]{article}

\usepackage{iccv}
\usepackage{times}
\usepackage{epsfig}
\usepackage{graphicx}
\usepackage{amsmath}
\usepackage{amssymb}
\usepackage{amsfonts}
\usepackage{subcaption}
\usepackage{tabularx}
\usepackage{bbm}

\usepackage{mathtools}
\usepackage{multirow}
\makeatletter

\graphicspath{{images/}}

\def\BState{\State\hskip-\ALG@thistlm}

\newcommand{\comment}[1]{}
\newcommand{\parag}[1]{\vspace{-3mm}\paragraph{#1}}

\newcommand{\phantomone}{\phantom{a}}
\newcommand{\phantomtwo}{\phantom{aa}}

\newcommand{\ourS}[0]{\emph{Ours-SRU}}
\newcommand{\ourD}[0]{\emph{Ours-DRU}}
\newcommand{\refN}[0]{\emph{RefineNet}}
\newcommand{\recS}[0]{\emph{Rec-Simple}}
\newcommand{\recL}[0]{\emph{Rec-Last}}
\newcommand{\recM}[0]{\emph{Rec-Middle}}
\newcommand{\Unet}[0]{\emph{U-Net}}

\newcommand{\TourS}[0]{{Ours-SRU}}
\newcommand{\TourD}[0]{{Ours-DRU}}
\newcommand{\TrefN}[0]{{RefineNet}}
\newcommand{\TrecS}[0]{{Rec-Simple}}
\newcommand{\TrecL}[0]{{Rec-Last}}
\newcommand{\TrecM}[0]{{Rec-Middle}}
\newcommand{\TUnet}[0]{{U-Net}}

\newif\ifdraft
\draftfalse

\newcommand{\ww}[1]{\ifdraft {\color{blue}{#1}} \else {}\fi}

\newcommand{\ky}[1]{\ifdraft {\color{blue}{#1}} \else {}\fi}

\newcommand{\KY}[1]{\ifdraft {\color{blue}{\textbf{KY: #1}}} \else {}\fi}

\usepackage{balance}
\usepackage{rotating}
\usepackage{booktabs}

 \iccvfinalcopy 

\def\iccvPaperID{845} 

\ificcvfinal\fi
\begin{document}

\newcommand{\STAB}[1]{\begin{tabular}{@{}c@{}}#1\end{tabular}}

\title{Recurrent U-Net for Resource-Constrained Segmentation}

\author{Wei Wang\thanks{The first two authors contribute equally.} \and Kaicheng Yu\textsuperscript{*} \and Joachim Hugonot \and Pascal Fua \and Mathieu Salzmann \\
	CVLab, EPFL, 1015 Lausanne \\
{\tt\small \{wei.wang\; kaicheng.yu\; joachim.hugonot\; pascal.fua\; mathieu.salzmann\}@epfl.ch}
}

\maketitle


\begin{abstract}

 State-of-the-art segmentation methods rely on very deep networks that are not always easy to train without very large training datasets and tend to be relatively slow to run on standard GPUs. In this paper, we introduce a novel recurrent U-Net architecture that preserves the compactness of the original U-Net~\cite{Ronneberger15}, while substantially increasing its performance to the point where it outperforms the state of the art on several benchmarks. We will demonstrate its effectiveness for several tasks, including hand segmentation, retina vessel segmentation, and road segmentation. We also introduce a large-scale dataset for hand segmentation.

\end{abstract}

\vspace{-6mm}
\section{Introduction}

While recent semantic segmentation methods achieve impressive results~\cite{Chen18c,Lin17c,Long15a,Zhao17b}, they require very deep networks and their architectures tend to focus on high-resolution and large-scale datasets and to rely on pre-trained backbones. For instance, state-of-the-art models, such as Deeplab~\cite{Chen17a,Chen18c}, PSPnet~\cite{Zhao17b} and RefineNet~\cite{Lin17c}, use a ResNet101~\cite{He16} as their backbone. This results in high GPU memory usage and inference time, and makes them less than ideal for operation in power-limited environments where real-time performance is nevertheless required, such as when segmenting hands using the onboard resources of an Augmented Reality headset.  This has been addressed by architectures such as the ICNet~\cite{zhao2018icnet} at the cost of a substantial performance drop. Perhaps even more importantly, training very deep networks usually requires either massive amounts of training data or image statistics close to that of ImageNet~\cite{Deng09}, which may not be appropriate in fields such as biomedical image segmentation where the more compact U-Net architecture remains prevalent~\cite{Ronneberger15}.

\begin{figure}[t]
	\includegraphics[width=\linewidth]{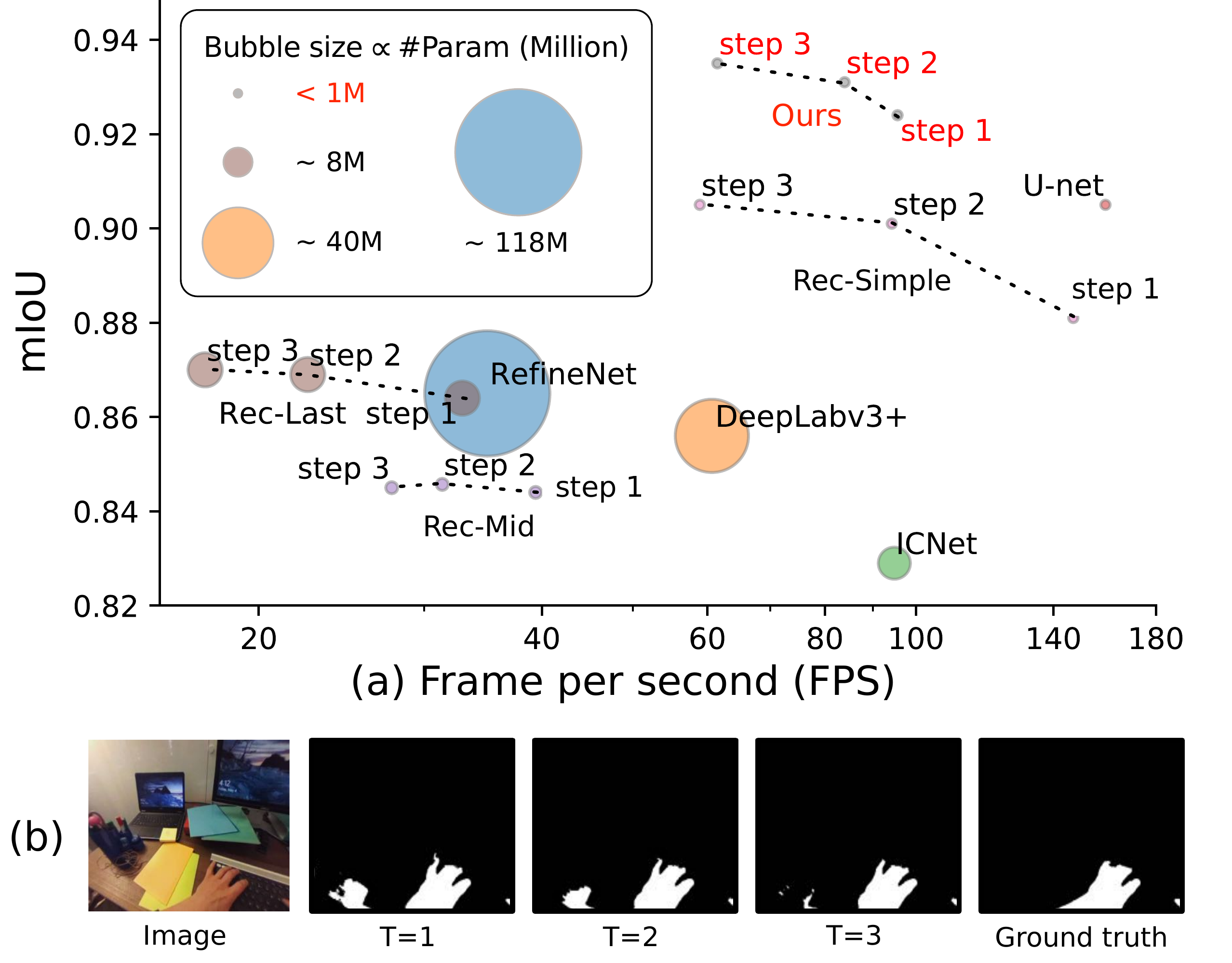}
	\vspace{-6mm}
	\caption{
		\small {
			\textbf{Speed vs accuracy.} Each circle represents the performance of a model in terms frames-per-second and mIoU accuracy on our Keyboard Hand Dataset using a Titan X~(Pascal) GPU. The radius of each circle denotes the models' number of parameters. For our recurrent approach, we plot these numbers after 1, 2, and 3 iterations, and we show the corresponding segmentations in the bottom row. The performance of our approach is plotted in red and the other acronyms are defined in Section~\ref{sec:baseline}. ICNet~\cite{zhao2018icnet} is slightly faster than us but at the cost of a significant accuracy drop, whereas RefineNet~\cite{Lin17c}  and DeepLab~\cite{Chen18c} are both slower and less accurate on this dataset, presumably because there are not enough training samples to learn their many parameters.}}
	\vspace{-2mm}
	\label{fig:teaser}
\end{figure}

In this paper, we argue that these state-of-the-art methods do not naturally generalize to resource-constrained situations and introduce a novel recurrent U-Net architecture that preserves the compactness of the original U-Net~\cite{Ronneberger15}, while substantially increasing its performance to the point where it outperforms the current state of the art on 5 hand-segmentation datasets, one of which is showcased in Fig.~\ref{fig:teaser}, 
and a retina vessel segmentation one. With only 0.3 million parameters, our model is much smaller than the ResNet101-based DeepLabv3+~\cite{Chen18c} and RefineNet~\cite{Lin17c}, with 40 and 118 million weights, respectively. This helps explain why we can outperform state-of-the-art networks on speclalized tasks: The pre-trained ImageNet features are not necessarily the best  and training sets are not quite as large as CityScapes~\cite{Cityscapes}. As a result, the large networks tend to overfit and do not perform as well as compact models trained from scratch. 

\begin{figure}[t]
	\begin{center}
		\includegraphics[width=\linewidth]{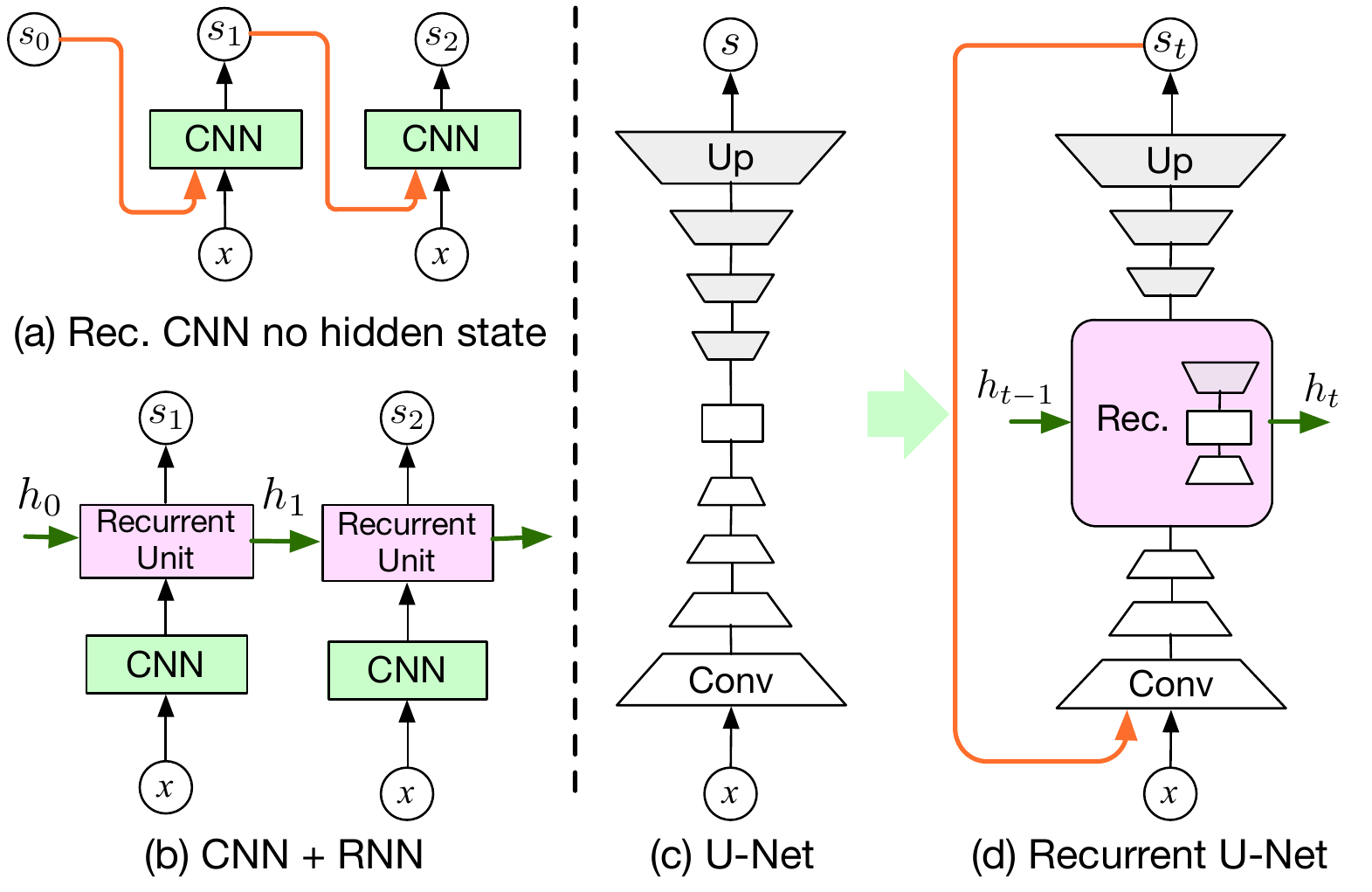}
	\end{center}
	\vspace{-3mm}
	\caption{{\bf Recurrent segmentation.} {\bf (a)} The simple strategy of~\cite{Mosinska18,Pinheiro14} consists of concatenating the previous segmentation mask $s_{t-1}$ to the image $x$, and recurrently feeding this to the network. {\bf (b)} For sequence segmentation, to account for the network's internal state, one can instead combine the CNN with a standard recurrent unit as in~\cite{Valipour17}. Here, we build upon the U-Net architecture of~\cite{Ronneberger15} {\bf  (c)}, and propose to build a recurrent unit over several of its layers, as shown in {\bf (d)}. This allows us to propagate higher-level information through the recurrence, and, in conjunction with a recurrence on the segmentation mask, outperforms the two simpler recurrent architectures  {\bf (a)} and {\bf (b)}.
	}
	\vspace{-5mm}
	\label{fig:rcnn}
\end{figure}

The standard  U-Net takes the image as input, processes it, and directly returns an output. By contrast, our recurrent architecture iteratively refines both the segmentation mask and the network's internal state. This mimics human perception as in the influential AutoContext paper~\cite{Tu09}:\KY{do we want to say 'influential' here?} When we observe a scene, our eyes undergo saccadic movements, and we accumulate knowledge about the scene and continuously refine our perception~\cite{Purves01}. To this end we retain the overall structure of the U-Net, but build a recurrent unit over some of its inner layers for internal state update. By contrast with the simple CNN+RNN architecture of Fig.~\ref{fig:rcnn}(b), often used for video or volumetric segmentation~\cite{Valipour17,Poudel16,Ballas16}, this enables the network to keep track of and to iteratively update more than just a single-layer internal state. This gives us the flexibility to choose the portion of the internal state that we exploit for recursion purposes and to explore variations of our scheme. 

We demonstrate the benefits of our recurrent U-Net on several tasks, including hand segmentation, retina vessel segmentation and road segmentation. Our approach consistently outperforms earlier and simpler approaches to recursive segmentation~\cite{Mosinska18,Poudel16,Valipour17}. For retina vessel segmentation, it also outperforms the state-of-the-art method of~\cite{maninis2016deep} on the DRIVE \cite{staal:2004-855} dataset, and for hand segmentation, the state-of-the-art RefinetNet-based method of~\cite{Urooj18} on several modern benchmarks~\cite{Fathi11,Bambach15,Urooj18}. 
As these publicly available hand segmentation datasets are relatively small, with at most 4.8K annotated images, we demonstrate the scalability of our approach, along with its applicability in a keyboard typing scenario, by introducing a larger dataset containing 12.5K annotated images. It is the one we used to produce the results shown in Fig.~\ref{fig:teaser}. We will make it publicly available along with our code upon acceptance of the paper.

Our contribution is therefore an effective recurrent approach to semantic segmentation that can operate in environments where the amount of training data and computational power are limited. It does not require more memory than the standard U-Net thanks to parameter sharing and does not require training datasets as large as other state-of-the-art networks do. It is practical for real-time application, reaching 55 frames-per-second (fps) to segment $230 {\times} 306$ images on an NVIDIA TITAN X with 12G memory. Furthermore, as shown in Fig.~\ref{fig:teaser}, we can trade some accuracy for speed by reducing the number of iterations. Finally, while we focus on resource-constrained applications, our model can easily be made competitive on standard benchmarks such as Cityscapes by modifying its backbone architecture. We will show that replacing the U-Net encoder by a VGG16 backbone yields performance numbers comparable to the state of the art on this dataset. 


\section{Related Work}
\vspace{2mm}
\parag{Compact Semantic Segmentation Models.}

State-of-the-art semantic segmentation techniques~\cite{Chen18c,Lin17c,Long15a,Zhao17b} rely on very deep networks, which makes them ill-suited in resource-constrained scenarios, such as real-time applications and when there are only limited amounts of training data. In such cases, more compact networks are preferable. Such networks fall under two main categories. 

The first group features encoder-decoder architectures~\cite{Ronneberger15,poudel2019fast,Badrinarayanan15,romera2018erfnet,pohlen2017full,saxena2016convolutional,fourure2017residual}. Among those, U-Net~\cite{Ronneberger15} has demonstrated its effectiveness and versatility on many tasks, in particular for biomedical image analysis where it remains a favorite. For example, a U-net like architecture was recently used to implement the flood-filling networks of~\cite{Januszewski18} and to segment densely interwoven neurons and neurites in teravoxel-scale 3D electron-microscopy image stacks. This work took advantage of the immense amount of computing power that Google can muster but, even then, it is unlikely that this could have been accomplished with much heavier architectures. 

The second type involves multi-branch structures~\cite{poudel2018contextnet,yu2018bisenet,zhao2018icnet} to fuse low-level and high-level features at different resolutions. These require careful design to balance speed against performance. By contrast, the U-Net relies on simpler skip connections and, thus, does not require a specific design, which has greatly contributed to its popularity. 

\parag{Recurrent Networks for Segmentation.}

The idea of recurrent segmentation predates the deep learning era and was first proposed in AutoContext~\cite{Tu09}, and recurrent random forest~\cite{shotton2008semantic}. It has inspired many recent approaches, including several that rely on deep networks. For example, in~\cite{Mosinska18}, the segmentation mask produced by a modified U-Net was passed back as input to it along with the original image, which resulted in a progressive refinement of the segmentation mask. Fig.~\ref{fig:rcnn}(a) illustrates this approach. A similar one was followed in the earlier work of~\cite{Pinheiro14}, where the resolution of the input image patch varied across the iterations of the refinement process.

Instead of including the entire network in the recursive procedure, a standard recurrent unit can be added at the output of the segmentation network, as shown in Fig.~\ref{fig:rcnn}(b). This was done in~\cite{Romera16} to iteratively produce individual segmentation masks for scene objects. In principle, such a convolutional recurrent unit~\cite{Ballas16,Poudel16,Valipour17} could also be applied for iterative segmentation of a single object and we will evaluate this approach in our experiments. We depart from this strategy by introducing gated recurrent units that encompass several U-Net layers. Furthermore, we leverage the previous segmentation results as input, not just the same image at every iteration.

Iterative refinement has also been used for pose estimation~\cite{ramakrishna2014pose,wei2016convolutional,newell2016stacked}. The resulting methods all involve consecutive modules to refine the predictions with a loss function evaluated on the output of each module, which makes them similar in spirit to the model depicted by Fig.~\ref{fig:rcnn}(a). Unlike in our approach, these methods do not share the parameters across the consecutive modules, thus requiring more parameters and moving away from our aim to obtain a compact network. Furthermore, they do not involve RNN-inspired memory units to track the internal hidden state.

\section{Method}
\vspace{-1mm}
We now introduce our novel recurrent semantic segmentation architecture. To this end, we first discuss the overall structure of our 
framework, and then provide the details of the recurrent unit it relies on. Finally, we briefly discuss the training strategy for our approach.

\subsection{Recurrent U-Net}
\vspace{-0mm}
\label{sec:r-unet}

\begin{figure*}[t]
	\vspace{-3mm}
	\includegraphics[width=\textwidth]{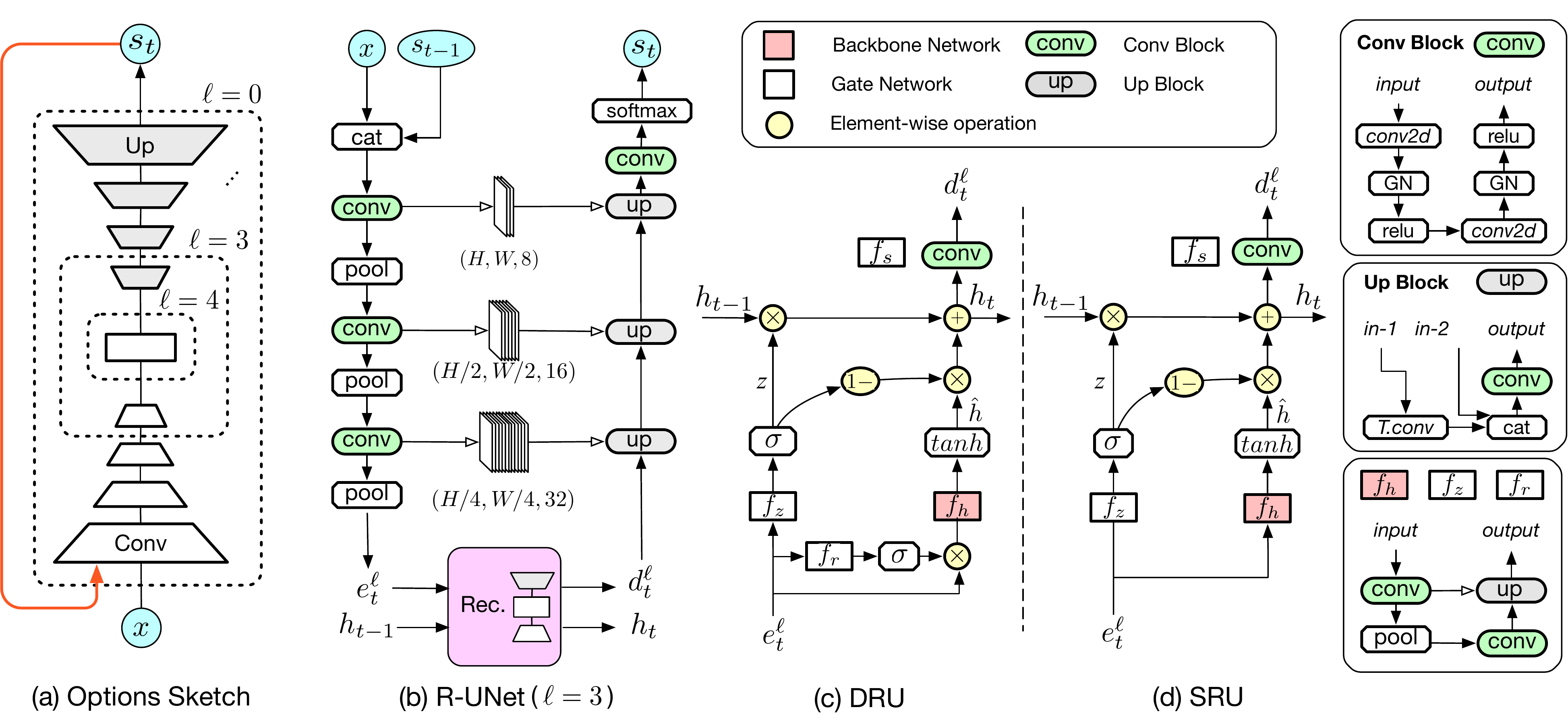}
	\vspace{-6mm}
	\caption{\textbf{Recurrent UNet (R-UNet).} {\bf (a)} As illustrated in Fig.~\protect\ref{fig:rcnn}(d), our model incorporates several encoding and decoding layers in a recurrent unit. The choice of which layers to englobe is defined by the parameter $\ell$. {\bf (b)} For $\ell = 3$, the recurrence occurs after the third pooling layer in the U-Net encoder. The output of the recurrent unit is then passed through three decoding up-convolution blocks. We design two different recurrent units, the Dual-gated Recurrent Unit (DRU)~\textbf{(c)} and the Single-gated Recurrent Unit (SRU)~\textbf{(d)}. They differ by the fact that the first one has an additional reset gate acting on its input. See the main text for more detail. }
	\label{fig:model_arch}
	\vspace{-1mm}
\end{figure*}

We rely on the U-Net architecture of~\cite{Ronneberger15} as backbone to our approach. As shown in Fig.~\ref{fig:model_arch}(a), the U-Net has an encoder-decoder structure, with skip connections between the corresponding encoding and decoding layers that allow the network to retain low-level features for the final prediction. Our goal being to operate in resource-constrained environments, we want to keep the model relatively simple. We therefore rely on a U-Net design where the first convolutional unit has 8 feature channels, and, following the original U-Net strategy, the channel number doubles after every pooling layer in the encoder. The decoder relies on transposed convolutions to increase the model's representation power compared to bilinear interpolation. We use group-normalization~\cite{Wu18a} in all convolutional layers since we usually rely on very small batch sizes.

Our contributions are to integrate recursions on 1) the predicted segmentation mask $s$ and 2) multiple internal states of the network. \ky{which is inspired from  hidden state of recurrent neural network.} \KY{add to further distinguish the 'activation as internal state'.} The former can be achieved by simply concatenating, at each recurrent iteration $t$, the previous segmentation mask $s_{t-1}$ to the input image, and passing the resulting concatenated tensor through the network. For the latter, we propose to replace a subset of the encoding and decoding layers of the U-Net with a recurrent unit. Below, we first formalize this unit, and then discuss two variants of its internal mechanism. 

To formalize our recurrent unit, let us consider the process at iteration $t$ of the recurrence. At this point, the network takes as input an image $x$ concatenated with the previously-predicted segmentation mask $s_{t-1}$. Let us then denote by $e^{\ell}_t$ the activations of the $\ell^{th}$ encoding layer, and by $d^{\ell}_t$ those of the corresponding decoding layer. Our recurrent unit takes as input $e^{\ell}_t$, together with its own previous hidden tensor $h_{t-1}$, and outputs the corresponding activations $d^{\ell}_t$, along with the new hidden tensor $h_t$. Note that, to mimic the computation of the U-Net, we use multiple encoding and decoding layers within the recurrent unit.

In practice, one can choose the specific level $\ell$ at which the recurrent unit kicks in. In Fig.~\ref{fig:model_arch}~(b), we illustrate the whole process for $\ell=3$. 
When $\ell=0$, the entire U-Net is included in the recurrent unit, which then takes the concatenation of the segmentation mask and the image as input. Note that, for $\ell = 4$, the recurrent unit still contains several layers because the central portion of the U-Net in Fig.~\ref{fig:model_arch}(a) corresponds to a convolutional {\it block}.
In our experiments, we evaluate two different structures for the recurrent units, which we discuss below.

\subsection{Dual-gated Recurrent Unit}
\label{sec:dru}
\vspace{-0mm}
As a first recurrent architecture, we draw inspiration from the Gated Recurrent Unit (GRU)~\cite{Cho14c}. As noted above, however, our recurrent unit replaces multiple encoding and decoding layers of the segmentation network. We therefore modify the equations accordingly, but preserve the underlying motivation of GRUs. Our architecture is shown in Fig.~\ref{fig:model_arch}(c).

Specifically, at iteration $t$, given the activations $e^{\ell}_t$ and the previous hidden state $h_{t-1}$, we aim to produce a candidate update $\hat{h}$ for the hidden state and combine it with the previous one according to how reliable the different elements of this previous hidden state tensor are. To determine this reliability, we use an update gate defined by a tensor 
\begin{equation}
z = \sigma( f_{z}(e^{\ell}_{t}) )\;,
\label{eq:p}
\end{equation}
where $f_{z}(\cdot)$ denotes an encoder-decoder network with the same architecture as the portion of the U-Net that we replace with our recurrent unit.

Similarly, we obtain the candidate update as
\begin{equation}
\hat{h} = tanh(f_{h}(r\odot e^{\ell}_{t}))\;,
\label{eq:q}
\end{equation}
where $f_{h}(\cdot)$ is a network with the same architecture as $f_{z}(\cdot)$, but a separate set of parameters, $\odot$ denotes the element-wise product, and $r$ is a reset tensor allowing us to mask parts of the input used to compute $\hat{h}$. It is computed as
\begin{equation}
r = \sigma( f_{r}(e^{\ell}_{t}) )\;,
\end{equation}
where $f_r(\cdot)$ is again a network with the same encoder-decoder architecture as before.

Given these different tensors, the new hidden state is computed as
\begin{equation}
h_t = z \odot h_{t-1} + (1-z) \odot \hat{h}\;.
\label{eq:ht}
\end{equation}
Finally, we predict the output of the recurrent unit, which corresponds to the activations of the $\ell^{th}$ decoding layer as
\begin{equation}
d^{\ell}_{t} = f_{s}(h_{t})\;,
\label{eq:st}
\end{equation}
where, as shown in Fig.~\ref{fig:model_arch}(c), $f_s(\cdot)$ is a simple convolutional block.
Since it relies on two gates, $r$ and $z$, we dub this recurrent architecture Dual-gated Recurrent Unit (DRU). 
One main difference with GRUs is the fact that we use multi-layer encoder-decoder networks in the inner operations instead of simple linear layers. Furthermore, in contrast to GRUs, we do not directly make use of the hidden state $h_{t-1}$ in these inner computations. This allows us not to have to increase the number of channels in the encoding and decoding layers compared to the original U-Net. Nevertheless, the hidden state is indirectly employed, since, via the recursion, $e^{\ell}_t$ depends on $d^{\ell}_{t-1}$, which is computed from $h_{t-1}$.

\subsection{Single-Gated Recurrent Unit}
\label{sec:sru}
\vspace{-1mm}
As evidenced by our experiments, the DRU described above is effective at iteratively refining a segmentation. However, it suffers from the drawback that it incorporates three encoder-decoder networks, which may become memory-intensive depending on the choice of $\ell$. To decrease this cost, we therefore introduce a simplified recurrent unit, which relies on a single gate, thus dubbed Single-gated Recurrent Unit (SRU).

Specifically, as illustrated in Fig.~\ref{fig:model_arch}(d), our SRU has a structure similar to that of the DRU, but without the reset tensor $r$. As such, the equations remain mostly the same as above, with the exception of the candidate hidden state, which we now express as
\begin{equation}
\hat{h} = tanh(f_{h}(e^{\ell}_{t}))\;.
\label{eq:q2}
\end{equation}
This simple modification allows us to remove one of the encoder-decoder networks from the recurrent unit, which, as shown by our results, comes at very little loss in segmentation accuracy.
\vspace{-2mm}

\subsection{Training}
\label{sec:loss}
To train our recurrent U-Net, we use the cross-entropy loss. More specifically, we introduce supervision at each iteration of the recurrence. To this end, we write our overall loss as
\begin{equation}
L = \sum_{t=1}^N w_t L_t,
\end{equation}
where $N$ represents the number of recursions, set to 3 in this paper, and $L_t$ denotes the cross-entropy loss at iteration $t$, which is weighted by $w_t$. 
\begin{equation}
	w_t = \alpha^{N-t}.
	\label{eq:rel_weight}
\end{equation}
The weight, by setting $\alpha \leq1$, increases monotonically with the iterations.  
In our experiments, we either set $\alpha = 1$, so that all iterations have equal importance, or $\alpha = 0.4$, thus encoding the intuition that we seek to put more emphasis on the final prediction. A study of the influence of $\alpha$ is provided in supplementary material, 
where we also discuss our training protocol in detail.


\section{Experiments}
\label{sec:exp}

\vspace{-1mm}
We compare the two versions of our Recurrent U-Net against the state of the art on several tasks including hand segmentation, retina vessel segmentation and road delineation. The hyper-parameters of our models were obtained by validation, as discussed in the supplementary material. We further demonstrate that the core idea behind our idea also applies to non-resource-constrained scenarios, such as Cityscapes, by increasing the size of the U-Net encoder.

\subsection{Datasets.}
\label{sec:datasets}
\vspace{2mm}

\parag{Hands.}

\begin{figure}[!tb]
	\vspace{-1mm}
	\includegraphics[width=\linewidth]{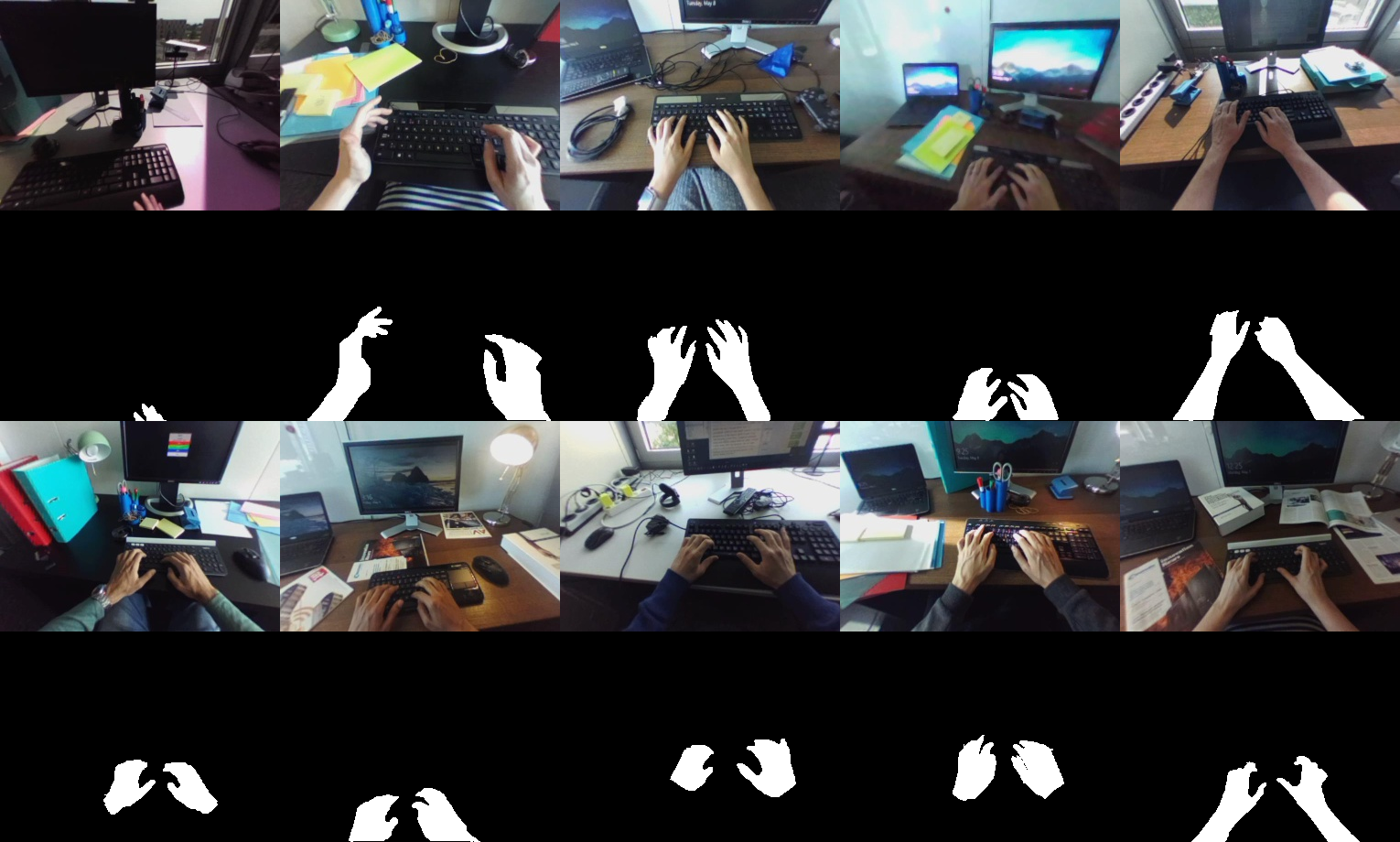}
	\vspace{-6mm}
	\caption{\small \textbf{Keyboard Hand (KBH) dataset.} Sample images featuring diverse environmental and lighting conditions, along with associated ground-truth segmentations.}
	\label{fig:annotEx}
	\vspace{-0mm}
\end{figure}

We report the performance of our approach on standard hand-segmentation benchmarks, such as GTEA~\cite{Fathi11}, EYTH~\cite{Urooj18}, EgoHand~\cite{Bambach15}, and HOF~\cite{Urooj18}. These, however, are relatively small, with at most 4,800 images in total, as can be seen in Table~\ref{tab:dsComp}. To evaluate our approach on a larger  dataset, we therefore acquired our own. Because this work was initially motivated by an augmented virtuality project whose goal is to allow someone to type on a keyboard while wearing a head-mounted display, we asked 50 people to type on 9 keyboards while wearing an HTC Vive~\cite{HTCvive}. To make this easier, we created a mixed-reality application to allow the users to see both the camera view and a virtual browser showing the text being typed.  To ensure diversity, we varied the keyboard types, lighting conditions, desk colors, and objects lying on them, as can be seen in Fig.~\ref{fig:annotEx}.  We provide additional details in Table~\ref{tab:kbh}.

We then recorded 161 hand sequences with the device's camera.   We split them as 20/ 20/ 60\% for train/ validation/ test to set up a challenging scenario in which the training data is not overabundant and to test the scalability and generalizability of the trained models. We guaranteed that the same person never appears in more than one of these splits by using people's IDs during partitioning. In other words, our splits resulted in three groups of 30, 30, and 101 separate videos, respectively. 
We annotated about the same number of frames in each one of the videos, resulting in a total of 12,536 annotated frames. 

\begin{table}[!t]
	\centering
	\resizebox{\linewidth}{!}{%
		\setlength{\tabcolsep}{4pt}
		\begin{tabular}[width=\textwidth]{l c  @{}cc c @{}rrrr  c}
			\toprule
			& \phantomone & \multicolumn{2}{c}{Resolution} & \phantomone& \multicolumn{4}{c}{\# Images} \\
			\cmidrule(r){3-4}  	\cmidrule(r){5-9} 
			Dataset && Width & Height &&   Train & Val. & Test & Total \\
			
			\midrule 
			%
			KBH (Ours)&& 230 & 306 && 2300 & 2300 & 7936 & 12536 \\
			\midrule 
			EYTH~\cite{Urooj18} && 216  & 384 && 774 & 258 & 258 &1290 \\
			HOF~\cite{Urooj18} && 216 & 384 && 198&40&62 & 300 \\
			EgoHand~\cite{Bambach15} && 720&1280 && 3600&400&800 & 4800 \\
			GTEA\cite{Fathi11} && 405&720 && 367 & 92 & 204 & 663\\
			\bottomrule
		\end{tabular}
	}
	\vspace{-1mm}
	\caption{\small Hand-segmentation benchmark datasets.}
	\label{tab:dsComp}
	\vspace{-2mm}
\end{table}

\begin{table}[!t]
	\setlength{\tabcolsep}{2pt}
	\resizebox{\linewidth}{!}{
		\begin{subtable}{0.5\linewidth}
			\centering
			\label{tab:ds_params}
			\resizebox{!}{45pt}{%
				\begin{tabular}{ccc}
					\multicolumn{3}{c}{(a) Environment setup}\\
					\toprule
					Parameters & Amount & Details \\
					\midrule
					Desk & 3 & White, Brown, Black\\
					Desk position & 3 & - \\
					Keyboard & 9 & -\\
					Lighting & 8 & 3 sources on/off\\
					Objects on desk & 3 & 3 different objects\\
					\bottomrule
				\end{tabular}
			}
		\end{subtable}%
		\begin{subtable}{.3\linewidth}
			\resizebox{!}{45pt}{%
				\begin{tabular}{c}
					\phantomtwo \\
				\end{tabular}%
			}
		\end{subtable} 
		\begin{subtable}{0.25\linewidth}
			\centering
			\label{tab:ds_attrib}
			\resizebox{!}{45pt}{%
				\begin{tabular}{cc}
					\multicolumn{2}{c}{(b) Attributes} \\
					\toprule
					Attribute & \#IDs \\
					\midrule
					Bracelet & 10\\
					Watch & 14\\
					Brown-skin & 2\\
					Tatoo & 1\\
					Nail-polish & 1\\
					Ring(s) & 6\\
					\bottomrule
				\end{tabular}%
			}
		\end{subtable} 
	}
	\vspace{-1mm}
	\caption{\small Properties of our new KBH dataset.}
	\label{tab:kbh}
	\vspace{-3mm}
\end{table}

\begin{figure*}
	\centering
	\includegraphics[width=0.95\linewidth, height=8cm]{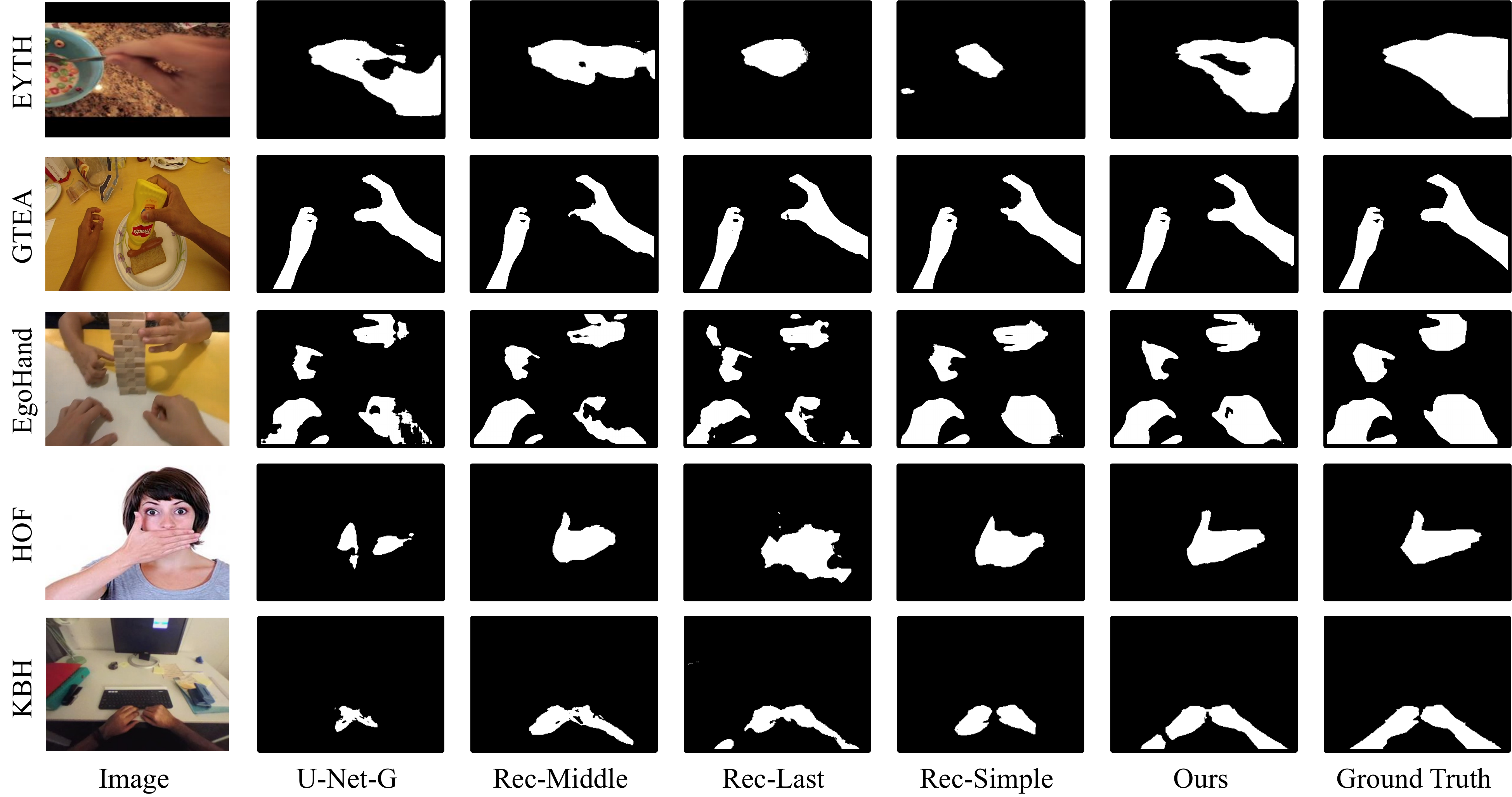}
	\vspace{-2mm}
	\caption{\small {\bf Example predictions on hand segmentation datasets.} 
		Note that our method yields accurate segmentations in diverse conditions, such as with hands close to the camera, multiple hands, hands over other skin regions, and low contrast images in our KBH dataset. By contrast, the baselines all fail in at least one of these scenarios. Interestingly,  our method sometimes yields a seemingly a more accurate segmentation than the ground-truth ones. For example, in our EYTH result at the top, the gap between the thumb and index finger is correctly found whereas it is missing from the ground truth. Likewise, for KBH at the bottom, the watch band is correctly identified as not being part of the arm even though it is labeled as such in the ground truth. 	
	}
	\label{fig:hand_vis}
	\vspace{-2mm}
\end{figure*}

\vspace{-1mm}
\parag{Retina Vessels.}
We used the popular DRIVE dataset~\cite{staal:2004-855}. It contains 40 retina images used for making clinical diagnoses, among which 33 do not show any sign of diabetic retinopathy and 7 show signs of mild early diabetic retinopathy. The images have been divided into a training and a test set with 20 images for each set.
\vspace{-1mm}
\parag{Roads.}
We used the Massachusetts Roads dataset~\cite{Mnih13}. It is one of the largest publicly available collections of aerial road images, containing both urban and rural neighborhoods, with many different kinds of roads ranging from small paths to highways. The data is split into 1108 training and 49 test images, one of which is shown in Fig.~\ref{fig:recSteps}.

\vspace{-1mm}
\parag{Urban landscapes.}
We employed the recent Cityscapes dataset. It is a very challenging dataset with high-resolution $1024 {\times} 2048$ images.
It has 5,000 finely annotated images which are split into training/validation/test sets with 2975/500/1525 images.
30 classes are annotated, and 19 of them are used in training and testing.

\subsection{Experimental Setup}
\label{sec:baseline}
\vspace{2mm}
\parag{Baselines.} 

We refer to the versions of our approach that rely on the dual gated unit of Section~\ref{sec:dru} and the single gated unit of  Section~\ref{sec:sru} as \ourS{} and \ourD{}, respectively, with, e.g., \ourS(3) denoting the case where $\ell=3$ in Fig.~\ref{fig:model_arch}. We compare them against the state-of-the-art model for each task, i.e., \refN{}~\cite{Urooj18} for hand segmentation, \cite{maninis2016deep} for retina vessel segmentation and \cite{Mosinska18} for road delineation, the general purpose DeepLab V3+~\cite{Chen18c}, the real-time ICNet~\cite{zhao2018icnet}, and the following baselines.
\begin{itemize}
	
	\comment{
		\item  {\bf \refN{}}~\cite{Urooj18}. The current state-of-the-art hand segmentation method that relies on RefineNet. On the standard benchmarks, we directly report the numbers from~\cite{Urooj18}. For our new dataset, we adopt the same training procedure as in~\cite{Lin17c}, also employed in~\cite{Urooj18}, which uses the pre-trained ResNet-101 on the Pascal Human-parts dataset. 
	}
	
	\vspace{-1mm}
	\item {\bf \Unet-B} and {\bf \Unet-G}~\cite{Ronneberger15}. We treat our U-Net backbone by itself as a baseline. \Unet-B{} uses batch-normalization and \Unet-G{}  group-normalization. For a fair comparison, they, \ourS{}, \ourD{}, and the recurrent baselines introduced below all use the same parameter settings.
	
	\vspace{-1mm}
	\item {\bf \recL}. It has been proposed to add a recurrent unit after a convolutional segmentation network to process sequential data, such as video~\cite{Poudel16}. The corresponding \Unet-based architecture can be directly applied to segmentation by inputing the same image at all time steps, as shown in Fig.~\ref{fig:rcnn}(b). The output then evolves as the hidden state is updated. 
	
	
	\vspace{-1mm}
	\item {\bf \recM }. Similarly, the recurrent unit can replace the bottleneck between the U-Net encoder and decoder, instead  of being added at the end of the network. This has been demonstrated to handle volumetric data~\cite{Valipour17}. Here we test it for segmentation. The hidden state then is of the same size as the inner feature backbone, that is, 128 in our  experimental setup. 
	
	\vspace{-1mm}
	\item {\bf \recS}~\cite{Mosinska18}. We perform a recursive refinement process, that is, we  concatenate the segmentation mask with the input image and feed it into the network. Note that the original method of~\cite{Mosinska18} relies on a VGG-19 pre-trained on ImageNet~\cite{Simonyan15}, which is far larger than our \Unet. To make the comparison fair, we therefore implement this baseline with the same U-Net backbone as in our approach.
	
	
\end{itemize}
\vspace{-2mm}
\parag{Scaling Up using Pretrained Deep Networks as Encoder}

While our goal is resource-constrained segmentation, our method extends to the general setting. In this case, to further boost its performance, we replace the U-Net encoder with a pretrained VGG-16 backbone. This process is explained in the supplementary material. We refer to the corresponding models as U-Net-VGG16 and DRU-VGG16.


\vspace{-1mm}
\parag{Metrics.}
We report the mean intersection over union (mIoU), mean recall (mRec) and mean precision (mPrec). 

\begin{table*}[t]
	\centering
	\resizebox{\textwidth}{!}{%
		\setlength{\tabcolsep}{4pt}
		\begin{tabular}[width=\textwidth]{@{}c <{\enspace } @{}   l| c @{}ccc c @{}ccc  c @{}ccc c @{}ccc c @{}ccc}
			\toprule
			& Model &  & \multicolumn{3}{c}{EYTH~\cite{Urooj18}} & & \multicolumn{3}{c}{GTEA~\cite{Fathi11}}  & & \multicolumn{3}{c}{EgoHand~\cite{Bambach15}} && \multicolumn{3}{c}{HOF~\cite{Urooj18}} && \multicolumn{3}{c}{KBH} \\
			\cmidrule(r){2-2} 		\cmidrule(r){4-6}  		\cmidrule(r){8-10}   		\cmidrule(r){12-14}   		\cmidrule(r){16-18}   		\cmidrule(r){20-22}  
			& & & mIOU  & mRec  & mPrec & & mIOU  & mRec  & mPrec & & mIOU  & mRec  & mPrec & & mIOU  & mRec  & mPrec & & mIOU  & mRec  & mPrec \\
			\cmidrule{2-22}
			& \textit{No pre-train} \\
			\multirow{11}{*}{\rotatebox{90}{\enspace Light }}
			& ICNet \cite{zhao2018icnet} & & 0.731 & 0.915 &  0.764 & &  0.898 & 0.971 & 0.922 & & 0.872 & \textbf{0.925} & 0.931 & &  0.580 &0.801 & 0.628 &&  0.829 & 0.925 & 0.876\\
			& \TUnet-B~\cite{Ronneberger15} && 0.803 &0.912 & 0.830 &  & 0.950 & 0.973 & 0.975 & & 0.815 & 0.869 & 0.876 && 0.694 & \textbf{0.867} & 0.778 && 0.870 & 0.943 & 0.911 \\
			& \TUnet-G &&0.837 & 0.928 & 0.883 & & 0.952 & 0.977 & 0.980 & &0.837 & 0.895 & 0.899 && 0.621 & 0.741 & 0.712  && 0.905 & 0.949 &0.948 \\
			& \TrecM~\cite{Poudel16} && 0.827 &0.920&0.877 &   & 0.924 &\textbf{0.979} & 0.976 && 0.828 &0.894 &0.905 && 0.654 &0.733 & \textbf{0.796}& & 0.845 & 0.924 & 0.898 \\
			& \TrecL~\cite{Valipour17} & & 0.838 &0.920  &0.894 &&0.957& 0.975 & 0.980 &&0.831&0.906 &0.897&& 0.674& 0.807 & 0.752  && 0.870 & 0.930	 & 0.924  \\
			& \TrecS~\cite{Mosinska18} & & 0.827 & 0.918 & 0.864 & & 0.952 & 0.975 & 0.976 && 0.858 & 0.909 & 0.931 & & 0.693 & 0.833 & 0.704 & &  0.905 & 0.951 & 0.944  \\
			\cmidrule{2-22}
			&  {\textit{Ours at layer ($\ell$) }} & \\
			& \TourS (0)  & & 0.844 &0.924&0.890&& \textbf{0.960} & 0.976 & 0.981 & & 0.862 &0.913 & 0.932 & & \textbf{0.712} & 0.844 & 0.764 && 0.930 & 0.968 & 0.957\\
			& \TourS (3)  && 0.845 & \textbf{0.931} &0.891 && 	0.956 & 0.977 & \textbf{0.982} &&0.864 &0.913 & 0.933 && 0.699 & 0.864 & 0.773  & & 0.921 &0.964& 0.951 \\
			& \TourD (4) & & \textbf{0.849} & 	0.926 &\textbf{ 0.900} &  & 0.958 &0.978 & 0.977&  & \textbf{0.873 }& 0.924 & \textbf{0.935} & & 0.709 & 0.866 & 0.774 & &  \textbf{0.935} & \textbf{0.980} & \textbf{0.970}\\
			\midrule
			\midrule
			& \textit{With pretrain} \\
			\multirow{4}{*}{\rotatebox{90}{\enspace Heavy}} 
			& \TrefN~\cite{Urooj18} & & 0.688 & 0.776 &  0.853 & &  0.821 & 0.869 & 0.928 & &  0.814 & 0.919 & 0.879 & &  0.766 & 0.882 &  0.859 &&  0.865 & 0.954 & 0.921 \\
			& Deeplab V3+ \cite{Chen18c} & & 0.757 & 0.819 &  0.875 & &  0.907 &0.928 & 0.976 & &  0.870 & 0.909 & \textbf{0.958} & &  0.722 &  0.822 &  0.816 &&  0.856 & 0.901 & 0.935\\
			& U-Net-VGG16 && 0.879 &0.945 & 0.921 &  & 0.961 & 0.978 & 0.981 & & 0.879 & 0.916 & 0.951 && 0.849 & 0.937 & 0.893 && 0.946 & 0.971 & 0.972 \\
			\cmidrule{2-22}
			& DRU-VGG16 && {0.897} & {0.946 } & {0.940} &  & \textbf{0.964} & \textbf{0.981} & \textbf{0.982} & & {0.892} & {0.925} & \textbf{0.958} && {0.863} & {0.948} & \textbf{0.901} && {0.954} & {0.973} & \textbf{0.979} \\
			\bottomrule
		\end{tabular}
	}
	\vspace{-1mm}
	\caption{\small \textbf{Comparing against the state of the art. }  According to the mIOU, \ourD(4) performs best on average, with \ourS(0) a close second. Generally speaking all recurrent methods do better than \refN{}, which represents the state of the art, on all datasets except HOF. We attribute this to HOF being too small for optimal performance without pre-training, as in \refN{}. This is confirmed by looking at DRU-VGG16, which yields the overall best results by relying on a pretrained deep backbone.
	}
	\label{tab:hand}
	\vspace{-0mm}
\end{table*}

\subsection{Comparison to the State of the Art}
\label{sec:SOTA}

We now compare the two versions of our approach to the state of the art and to the baselines introduced above on the tasks of hand segmentation, retina vessel segmentation and road delineation. We split the methods into the light ones and the heavy ones. The light models contain fewer parameters and are trained from scratch, whereas the heavy ones use a pretrained deep model as backbone. \ky{Furthermore, we provide a qualitative assessment of our model for recurrent refinement in Fig.~\ref{fig:recSteps}, more information about recurrent steps are discussed in supplementary materials. }

\parag{Hands.} 
As discussed in Section~\ref{sec:datasets}, we tested our approach using 4 publicly available datasets and our own large-scale one. We compare it against the baselines in Table~\ref{tab:hand} quantitatively and in Fig.~\ref{fig:hand_vis} qualitatively. 

Overall, among the light models, the recurrent methods usually outperform the one-shot ones, \emph{i.e}, ICNet~\cite{zhao2018icnet} and \Unet{}.
Besides, among the recurrent ones, \ourD(4) and \ourS(0) clearly dominate with \ourD(4) usually outperforming \ourS(0) by a small margin. Note that, even though \ourD(4)  as depicted by Fig.~\ref{fig:model_arch}(a)  looks superficially similar to \recM{}, they are quite different because \ourD{} takes the segmentation mask as input and relies on our new DRU gate, as discussed at the end of Section~\ref{sec:r-unet} and in Section~\ref{sec:dru}.  To confirm this, we evaluated a simplified version of \ourD(4) in which we removed the segmentation mask from the input. The validation mIOU on EYTH decreased from 0.836 to 0.826 but remained better than that of \recM{} which is 0.814.

Note that \ourD(4) is better than the heavy \refN{} model on 4 out of the 5 datasets, despite \refN{} representing the current state of the art. The exception is HOF, and we believe that this can be attributed to HOF being the smallest dataset, with only 198 training images. Under such conditions,  \refN{}  strongly benefits from exploiting a ResNet-101 backbone that was pre-trained on PASCAL person parts~\cite{Chen14a}, instead of training from scratch as we do. This intuition is confirmed by looking at the results of our DRU-VGG16 model, which, by using a pretrained deep backbone, yields the overall best performance.


\paragraph{Model Performance, Size and Speed.}

\begin{figure}[t]
	\centering
	\resizebox{\linewidth}{!}{
		\includegraphics[width=\linewidth]{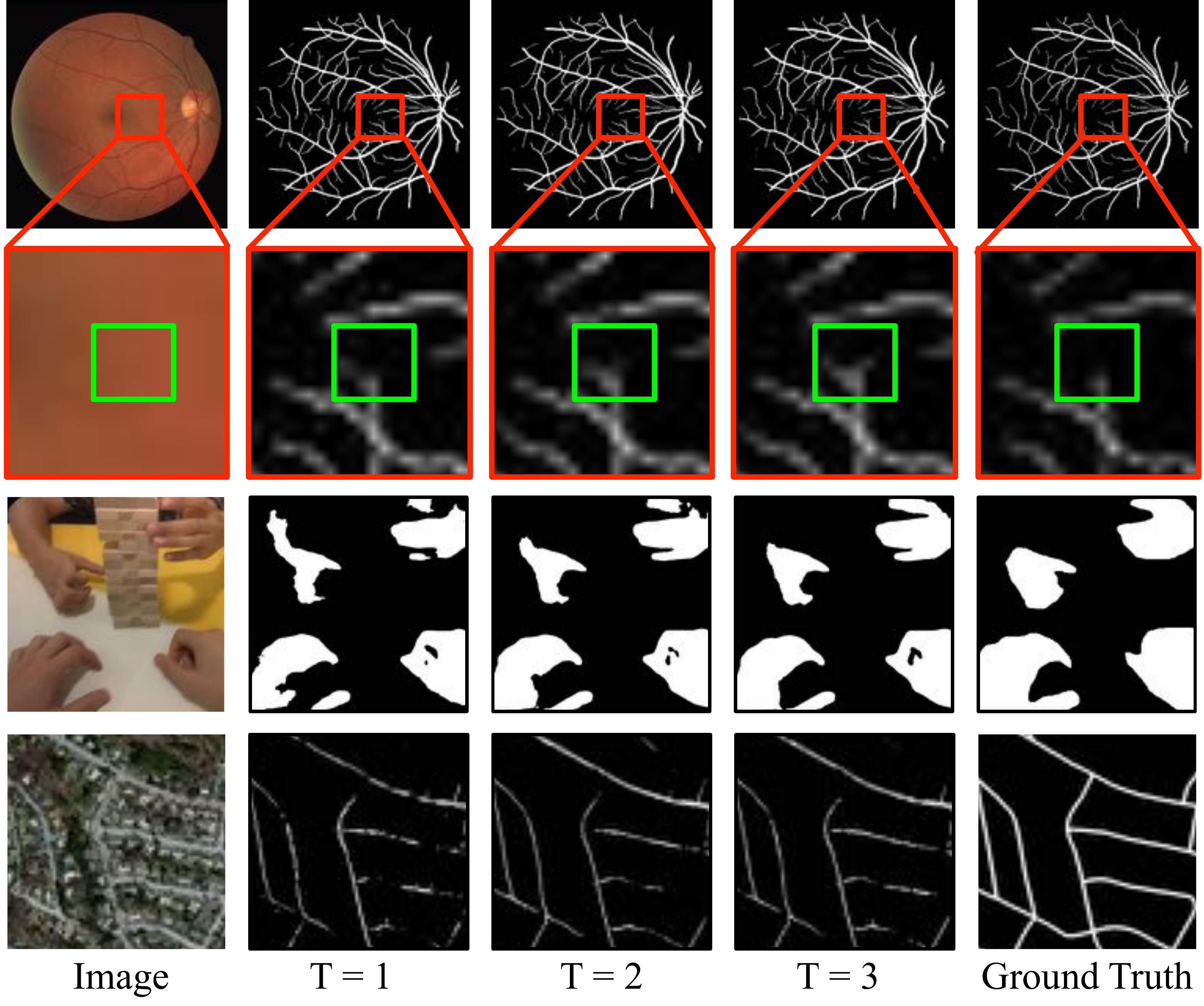}
		\vspace{-1mm}
	}
	\caption{
		\small {\bf Recursive refinement.} Retina, hand and road images; segmentation results after 1, 2, and 3 iterations; ground truth. Note the progressive refinement and the holes of the vessels, hands and roads being filled recursively. It is worth pointing out that even the tiny vessel branches in the retina which are ignored by the human annotators could be correctly segmented by our algorithm.  Better viewed in color and zoom in.}
	\vspace{-2mm}
	\label{fig:recSteps}
\end{figure}

Table~\ref{tab:hand} shows that DRU-VGG16 outperforms \ourD{}, e.g., by 0.02 mIoU points on KBH. This, however, comes at a cost.
To be precise, DRU-VGG16 has 41.38M parameters. This is 100 times larger than \ourD(4), which has only 0.36M parameters.
Moreover, DRU-VGG16 runs only at 18 fps, while \ourD(4) reaches 61 fps. This makes DRU-VGG16, and the other heavy models, ill-suited to embedded systems, such as a VR camera, while \ourD{} can more easily be exploited in resource-constrained environments.

\parag{Retina Vessels.}

\begin{table}[t]
	\centering
	\setlength{\tabcolsep}{8pt}
	\resizebox{ \linewidth}{!}{
		\begin{tabular}[width=\linewidth] {@{}c <{\enspace } @{} l | c ccccc @{}}
			\toprule
			&{Models} && mIOU & mRec & mPrec  & mF1 \\
			\cmidrule{2-7}
			\multirow{6}{*}{\rotatebox{90}{\enspace Light }}
			& ICNet \cite{zhao2018icnet} && 0.618 & 0.796 & 0.690 & 0.739 \\
			& \TUnet-G~\cite{Ronneberger15} && 0.800 & 0.897 &0.868 & 0.882 \\

			& \TrecM~\cite{Poudel16} && 0.818 & \textbf{0.903} & 0.886 & 0.894 \\
			& \TrecS~\cite{Mosinska18} && 0.814 & 0.898 & 0.885 & 0.892 \\
			& \TrecL~\cite{Valipour17} && 0.819 & 0.900 & 0.890 & 0.895 \\
			\cmidrule{2-7}
			
			& \TourD (4)  && \textbf{0.821} &0.902& \textbf{0.891} & \textbf{0.896} \\
\midrule
\midrule
			\multirow{3}{*}{\rotatebox{90}{\enspace Heavy  }}
			& DeepLab V3+ \cite{Chen18c} && 0.756 & 0.875 & 0.828 & 0.851 \\
			& U-Net-VGG16  && 0.804 & \textbf{0.910} & 0.862 & 0.886 \\
			& DRU-VGG16  && \textbf{0.817} & 0.905 & \textbf{0.883}  & \textbf{0.894} \\
			\bottomrule
		\end{tabular}
	}
	\vspace{-2mm}
	\caption{\small \textbf{Retina vessel segmentation results.}}
	\label{tab:drive}
	\vspace{-2mm}
\end{table}

We report our results in Table~\ref{tab:drive}. 
	Our DRU yields the best mIOU, mPrec and mF1 scores. Interestingly, on this dataset, it even outperforms the larger DRU-VGG16 and DeepLab V3+, which performs comparatively poorly on this task. This, we believe, is due to the availability of only limited data, which leads to overfitting for such a very deep network. Note also that retina images significantly differ from the ImageNet ones, thus reducing the impact of relying on pretrained backbones. On this dataset,~\cite{maninis2016deep} constitutes the state of the art, reporting an F1 score on the vessel class only of 0.822. According to this metric, \ourD(4) achieves 0.92, thus significantly outperforming the state of the art.
	
\comment{	
	Besides, the F1 score of the vessels is 0.920 for our DRU method, which is much larger than 0.780 for the U-Net baseline and 0.822 in \cite{maninis2016deep} which also emplys pretrained VGG network as the backbone.
	It is very interesting to observe that our original DRU network has better performance than DRU-VGG16.
	The DeepLab V3+ also has very poor performance.
	This observation is in contrast with the observation in the hand segmentation datasets. 
	One reason is that the retina images are very limited, and this will lead to overfitting for the very deep networks.
	Besides, the retina images are also visually quite different from the ones in ImageNet which are used to pretrain the deep networks.
	Thus, the pretrained parameters do not work so well on the retina images.
	}

\parag{Roads.}
Our results on road segmentation are provided in Table~\ref{tab:road}. We also outperform all the baselines by a clear margin on this task, with or without ImageNet pre-training.  In particular, \TourD (4) yields an mIoU 8 percentage point (pp) higher than \TUnet-G, and DRU-VGG16 5pp higher than U-Net-VGG16. This verifies that our recurrent strategy helps.
Furthermore, \TourD (4) also achieves a better performance than DeepLab V3+ and U-Net-VGG16.  Note that, here, we also report two additional metrics: Precision-recall breaking point~(P/R) and F1-score. The cutting threshold for all metrics is set to 0.5 except for P/R. For this experiment, we did not report the results of \Unet-B because  \Unet-G is consistently better. 

Note that a P/R value of 0.778 has been reported on this dataset in~\cite{Mosinska18}. However, this required using an additional topology-aware loss and a \Unet{} much larger than ours, that is, based on 3 layers of a VGG19 pre-trained on ImageNet. \TrecS{} duplicates the approach of~\cite{Mosinska18} without the topology-aware loss and with the same \Unet{} as  \TourD{}. Their mIoU of 0.723, inferior to ours of 0.757, shows our approach to recursion to be beneficial.


\begin{table}[t]
	\centering
	\setlength{\tabcolsep}{5pt}
	\resizebox{\linewidth}{!}{
		\begin{tabular}[width=\linewidth] {@{}c <{\enspace } @{} l | c @{}ccccc}
			\toprule
			& Models && mIOU & mRec & mPrec & P/R & mF1 \\
			\cmidrule{2-8}
			\multirow{6}{*}{\rotatebox{90}{\enspace Light }}
			& ICNet \cite{zhao2018icnet} && 0.476 & 0.626 & 0.500 & 0.513 & 0.656 \\
			& \TUnet-G~\cite{Ronneberger15} && 0.479 & 0.639 &0.502 & 0.642 & 0.563 \\
			&\TrecM~\cite{Poudel16} && 0.494 & 0.767 & 0.518 & 0.660 &  0.574 \\
			&\TrecS~\cite{Mosinska18} && 0.534 & 0.802 & 0.559 & 0.723 & 0.659 \\
			&\TrecL~\cite{Valipour17} && 0.526 & 0.786 & 0.551 & 0.730 & 0.648 \\
			\cmidrule{2-8}
			&\TourD (4)  && \textbf{0.560} &\textbf{ 0.865}& \textbf{0.583} & \textbf{0.757} & \textbf{0.691 } \\
			\midrule
			\midrule
			\multirow{3}{*}{\rotatebox{90}{\enspace Heavy  }}
			&Deeplab V3+ \cite{Chen18c} && 0.529 & 0.763 &0.555 &  0.710 & 0.643 \\
			&U-Net-VGG16  && 0.521 & 0.836 & 0.544 & 0.745 & 0.659 \\
           &DRU-VGG16  &&\textbf{0.571}& \textbf{0.862 }& \textbf{0.595} &\textbf{ 0.761 }& \textbf{0.704} \\
			\bottomrule
	\end{tabular}}
	\vspace{-2mm}
	\caption{ \textbf{Road segmentation results.}}
	\label{tab:road}
	\vspace{-0mm}
\end{table}

\parag{Urban landscapes.}

\begin{table}[!tb]
	\centering
	\setlength{\tabcolsep}{12pt}
	\resizebox{0.95\linewidth}{!}{
		\begin{tabular}[width=\linewidth] {lc | lc}
			\toprule
			Model        & mIoU  & Model       & mIoU  \\ 
			\midrule
			ICNet\cite{zhao2018icnet} &  0.695 & DeepLab V3 \cite{Chen17a} & 0.778\\ 
			U-Net-G      & 0.429 & U-Net-G $\times$2  & 0.476 \\ 
			Rec-Last     & 0.502 & Rec-Last $\times$2 & 0.521 \\ 
			DRU(4)       & 0.532 & DRU(4) $\times$2   & 0.627 \\ 
			 &  & DRU-VGG16   & 0.761 \\
			\bottomrule
		\end{tabular}
	}
	\vspace{-2mm}
	\caption{ \small \textbf{Cityscapes Validation Set with Resolution $1024{\times}2048$. } $\times$2 indicates that we doubled the number of channels in the U-Net backbone. Note that, for our method, we do not use multi-scaling or horizontal flips during inference.}
	\label{tab:cityscapes}
	\vspace{-0mm}
\end{table}

The segmentation results on the Cityscapes validation set are shown in Table~\ref{tab:cityscapes}.
Note that \ourD{} is consistently better than U-Net-G and than the best recurrent baseline, \emph{i.e.}, Rec-Last. Furthermore, 
doubling the number of channels of the U-Net backbone increases accuracy, and so does using a pretrained VGG-16 as encoder.
Ultimately, our DRU-VGG16 model yields comparable accuracy with the state-of-the-art DeepLab V3 one, despite its use of a ResNet101 backbone. 

\section{Conclusion}

We have introduced a novel recurrent \Unet{} architecture that preserves the compactness of the original one, while substantially increasing its performance. At its heart is the fact that the recurrent units encompass several encoding and decoding layers of the segmentation network.  In the supplementary material we demonstrate it running in real-time on a virtual reality device.  We also introduced a new hand segmentation dataset that is larger than existing ones. 

In future work, we will extend our approach of recurrent unit to other backbones than \Unet{} and to multi-scale recurrent architectures.

\balance
{\small
\bibliographystyle{ieee}
\bibliography{ms}
}
%

\end{document}


\end{document}